\pdfoutput=1
\documentclass[11pt]{article}

\usepackage{ACL2023}

\usepackage{times}
\usepackage{latexsym}

\usepackage[T1]{fontenc}
\usepackage{CJKutf8}

\usepackage[utf8]{inputenc}

\usepackage{microtype}

\usepackage{inconsolata}

\usepackage{tcolorbox}
\tcbuselibrary{listingsutf8}

\newtcolorbox{promptbox}[1][]{
  colback=gray!5,
  colframe=gray!40,
  fonttitle=\bfseries,
  sharp corners,
  boxrule=0.1mm,
  listing only,
  listing options={
    basicstyle=\ttfamily\footnotesize,
    breaklines=true,
    tabsize=2,
    keepspaces=true,
    columns=fullflexible,
    showstringspaces=false,
  },
  #1
}

\title{A Comparative Analysis of Word Segmentation, Part-of-Speech Tagging, and Named Entity Recognition for Historical Chinese Sources, 1900-1950}

\author{Zhao Fang \\
  \textnormal{University of Chicago} \\
  \textnormal{zhaofang@uchicago.edu}\\
  \And
  Liang-Chun Wu \\
  \textnormal{University of Chicago}\\
  \And
  Xuening Kong \\
  \textnormal{Purdue University}\\
  \And
  Spencer Dean Stewart \\
  \textnormal{Purdue University} \\
  \textnormal{stewa443@purdue.edu}\\
}

\begin{document}
\maketitle

\begin{abstract}
This paper compares large language models (LLMs) and traditional natural language processing (NLP) tools for performing word segmentation, part-of-speech (POS) tagging, and named entity recognition (NER) on Chinese texts from 1900 to 1950. Historical Chinese documents pose challenges for text analysis due to their logographic script, the absence of natural word boundaries, and significant linguistic changes. Using a sample dataset from the Shanghai Library Republican Journal corpus, traditional tools such as Jieba and spaCy are compared to LLMs, including GPT-4o, Claude 3.5, and the GLM series. The results show that LLMs outperform traditional methods in all metrics, albeit at considerably higher computational costs, highlighting a trade-off between accuracy and efficiency. Additionally, LLMs better handle genre-specific challenges such as poetry and temporal variations (i.e., pre-1920 versus post-1920 texts), demonstrating that their contextual learning capabilities can advance NLP approaches to historical texts by reducing the need for domain-specific training data.
\end{abstract}

\section{Introduction}

With the large-scale digitization of historical documents, researchers are increasingly interested in how Natural Language Processing (NLP) methods might be used and adapted to address the unique characteristics of older texts \cite{guldi2023dangerous, ehrmann2023named, manjavacas2022adapting, piotrowski2012natural}. Classification models for tasks such as Named Entity Recognition (NER) have improved significantly with the development of neural-based approaches. However, their precision for historical materials still lags behind that of models trained on contemporary texts \cite{ehrmann2023named}. Recent applications of language model-based approaches to NLP tasks have shown mixed results for using large language models (LLMs) such as ChatGPT to generate universal NER output \cite{qin2023chatgpt}, including for historical documents \cite{gonzalez2023yes}. More targeted, domain-specific approaches have also proven effective \cite{polak2024extracting}, including classification tasks common in digital humanities research \cite{bamman2024classification} and in low-resource settings \cite{frei2023annotated, wang2023gpt}.

The processing of historical Chinese documents presents unique challenges for NLP tasks due to the logographic writing system, the absence of natural word boundaries, and the rich morphological structures embedded within individual characters \cite{cui-etal-2020-revisiting}. Previous work on relatively “simple” tasks, such as Chinese word segmentation, has evolved through three paradigm shifts: rule-based systems, statistical machine learning models, and LLMs based on the transformer architecture \cite{fang2024methods}. Traditional machine learning methods such as Jieba and spaCy rely on dictionary matching and hidden Markov models to identify word boundaries. The dramatic linguistic and logographical transformations that occurred in China during the late nineteenth and twentieth centuries \cite{liu1995translingual, tsu2023kingdom} pose particular challenges for these models, which struggle to handle out-of-vocabulary terms. Some researchers have approached this problem by first converting historical sources into standardized simplified Chinese before performing NLP tasks \cite{stewart2025methodology}. Others have drawn from domain-specific approaches to manually curate datasets from historical sources to improve tasks such as segmentation \cite{luo2019pkuseg, blouin2023unlocking}.   

The advent of LLMs capable of detecting contextual patterns from large corpora presents new opportunities for processing classical and modern Chinese texts. Although there has been growing interest in BERT-based models and the development of domain-specific tools to process historical Chinese sources \cite{yu2020bert, cui-etal-2020-revisiting, fang2024methods}, further research is needed to evaluate LLMs' performance on NLP tasks. This short research paper presents a comparative analysis of machine learning and LLM-based tools for word segmentation, part-of-speech (POS) tagging, and NER on a diverse set of sample texts taken from the Shanghai Library Republican Journal corpus.\footnote{https://textual-optics-lab.uchicago.edu/shanghai-library-republican-journal-corpus} This study finds that, for transitional-era Chinese texts, LLM-based approaches outperform traditional NLP tools on segmentation, POS tagging, and NER tasks. However, these improvements come with notable increases in computational costs, highlighting a trade-off between performance and efficiency.

\section{Methodology}

To create our ground truth files, we extracted a random sample of passages from a large textual dataset of Late-Qing and Republican periodicals held by the Shanghai Library. We identified 208 passages spanning the decades 1900 to 1950. These passages include a variety of genres and topics, such as government reports, academic writing, social and political commentary, and literary texts such as short stories and poetry. To assess the ability of existing tools to handle different genres and textual changes over time, our sample included 41 passages identified as poetry, with the the remaining 167 passages distributed across five decades: 1900 (23 passages), 1910 (32 passages), 1920 (40 passages), 1930 (34 passages) and 1940 (38 passages). The passages ranged in length from 6 to 170 characters, with an average of 41.3 characters and a total of 8,610 characters. From this sample, the authors collectively segmented and tagged the passages, with each passage verified by two authors. Discrepancies were noted and resolved after further discussion. 

We selected widely used and reputable tools for Chinese segmentation and NER, as well as several popular LLMs, to evaluate out-of-the-box performance of these tools in our comparative analysis. To evaluate their effectiveness, we generated consistent prompts for each LLM and utilized their APIs to ensure standardized conditions. The prompts included clear, precise instructions requiring that the results be provided in a structured JSON format. This enabled a straightforward comparison with our established ground-truth dataset.

\begin{CJK}{UTF8}{bsmi}
\begin{promptbox}[title=LLM API Prompt]
\footnotesize
You are a spaCy-style NLP annotator for Traditional Chinese text from 1900-1950. 

Do not remove any text, including punctuation and brackets. Don't treat spaces as tokens.

Tasks:

1. Segment the input text into tokens.

2. Annotate each token with:

- text: the exact token string
    
- pos: a coarse POS tag (POS tags are exclusively: \{list of tags\})
    
- ent: the entity label if the token is part of a named entity (NER types are exclusively: \{list of tags\}, otherwise "")

3. Return the result as a JSON list of objects.

Here is an example of expected input and output:

input text = "此問題為英國政治上第一棘手之難問題。"

expected output = [

    \{"text": "此", "pos": "DET", "ent": ""\},
    
    \{"text": "問題", "pos": "NOUN", "ent": ""\},
    
    \{"text": "為", "pos": "VERB", "ent": ""\},
    
    \{"text": "英國", "pos": "PROPN", "ent": "GPE"\},
    
    \{"text": "政治", "pos": "NOUN", "ent": ""\},
    
    \{"text": "上", "pos": "ADP", "ent": ""\},
    
    \{"text": "第一", "pos": "NUM", "ent": ""\},
    
    \{"text": "棘手", "pos": "ADJ", "ent": ""\},
    
    \{"text": "之", "pos": "PART", "ent": ""\},
    
    \{"text": "難", "pos": "ADJ", "ent": ""\},
    
    \{"text": "問題", "pos": "NOUN", "ent": ""\},
    
    \{"text": "。", "pos": "PUNCT", "ent": ""\}
  
  ]

Nothing else but valid JSON in the final response.

\end{promptbox}
\end{CJK}

The performance of each approach was assessed based on several key metrics:
\begin{enumerate}
    \setlength{\itemsep}{0pt} 
    \setlength{\parskip}{0pt} 
    \item F1 Score: As the standard metric for evaluating Chinese tokenization, the F1 score effectively balances the risks of over-tokenization and under-tokenization. An F1 score of 90\% or higher is generally considered indicative of high accuracy.
    \item Part-of-Speech (POS) Accuracy (\%): This metric measures the accuracy of POS tagging for those tokens that were correctly segmented.
    \item Named Entity Recognition (NER) Accuracy (\%): This measures the precision of named entity tagging for those tokens that were correctly segmented.
    \item Time (in seconds): The processing speed for each approach was recorded to assess efficiency.
    \item Tokens Sent/Received (for LLM models only): For the LLMs, we tracked the number of tokens sent and received to capture resource usage and cost implications.
    \item Failed (for LLM models only): For the LLMs, we tracked how often they didn’t return the file in the proper JSON format. 
\end{enumerate}

These metrics help assess and compare the performance, accuracy, and efficiency of different Chinese NLP tools and LLMs.

\section{Findings}

\subsection{Global Results}

As summarized in Table 1, with one exception, the LLM models outperformed traditional NLP tools across all metrics. Among the traditional tools, spacy\_bert performed best. LLMs required considerably more computational resources compared to traditional NLP tools. While OpenAI’s o3-mini tended to outperform other models, this improvement came at the significant expense of both time and tokens. Of the LLMs, GPT-4o and Claude-3.5-sonnet performed best in balancing high accuracy and speed, while Claude-3.5-haiku struggled to return the output in the proper format. The GLM-4 models scored the lowest on these tests, with NER accuracy for GLM-4-Long being lower than that of traditional models.

\begin{table*}[ht]
\centering
\resizebox{\textwidth}{!}{%
\begin{tabular}{lcccrrrr}
\hline
\textbf{Model} & \textbf{F1\_Score (\%)} & \textbf{POS\_Accuracy (\%)} & \textbf{NER\_Accuracy (\%)} & \textbf{Time (s)} & \textbf{Token\_Sent} & \textbf{Token\_Received} & \textbf{Failed} \\
\hline
jieba             & 81.72 & 42.07 & 93.74 & 7.57  & - & - & - \\
spacy\_jieba\_sm  & 82.14 & 67.13 & 92.35 & 1.49  & - & - & - \\
spacy\_jieba\_lg  & 82.14 & 72.56 & 92.96 & 1.68  & - & - & - \\
spacy\_default\_sm & 82.50 & 69.79 & 91.92 & 2.36  & - & - & - \\
spacy\_default\_lg & 82.50 & 73.74 & 93.28 & 1.98  & - & - & - \\
spacy\_bert       & 82.50 & 78.36 & 93.78 & 32.08 & - & - & - \\
\hline
gpt-4o                      & 91.97 & 86.28 & 96.40 & 796.61  & 111220 & 102764 & 0  \\
gpt-4o-mini-2024-07-18       & 90.98 & 84.01 & 96.26 & 1703.89 & 111220 & 104401 & 1  \\
o3-mini-2025-01-31          & 94.50 & 88.83 & 97.00 & 5295.68 & 111012 & 709125 & 0  \\
claude-3-5-sonnet-20241022  & 93.41 & 87.35 & 94.24 & 1485.59 & 130994 & 122294 & 1  \\
claude-3-5-haiku-20241022   & 86.59 & 86.29 & 95.25 & 1639.79 & 130994 & 121525 & 13 \\
GLM-4-0520                 & 88.30 & 83.94 & 95.31 & 3301.15 & 110223 & 101494 & 5  \\
GLM-4-Long                 & 89.54 & 83.49 & 90.62 & 2411.52 & 108730 & 107873 & 0  \\
\hline
\end{tabular}%
}
\caption{Results for Segmentation Accuracy, POS Accuracy, NER Accuracy, Processing Time, Tokens, and Failed Returns.}
\label{tab:model_comparison}
\end{table*}

\subsection{Poetry vs. Non-Poetry}

As found in Table 2, when comparing the performance of poetry versus non-poetry texts, traditional models consistently performed better on non-poetry texts with the exception of NER. Among LLMs, the differences between poetry and non-poetry were less pronounced, indicating the ability of LLMs to handle a greater variety of texts such as poetry. 

\begin{table*}[ht]
\centering
\resizebox{\textwidth}{!}{%
\begin{tabular}{lcccccc}
\hline
 & \multicolumn{2}{c}{\textbf{Seg\_F1 (\%)}} & \multicolumn{2}{c}{\textbf{POS\_Accuracy (\%)}} & \multicolumn{2}{c}{\textbf{NER\_Accuracy (\%)}} \\
\textbf{Model} & \textbf{Non-Poetry} & \textbf{Poetry} & \textbf{Non-Poetry} & \textbf{Poetry} & \textbf{Non-Poetry} & \textbf{Poetry} \\
\hline
jieba             & 84.43 & 70.71 & 46.65 & 23.64 & 93.03 & 96.61 \\
spacy\_jieba\_sm  & 84.71 & 71.65 & 70.13 & 54.94 & 91.59 & 95.43 \\
spacy\_jieba\_lg  & 84.71 & 71.65 & 76.23 & 57.62 & 92.68 & 94.08 \\
spacy\_default\_sm & 85.28 & 71.19 & 72.91 & 57.09 & 91.19 & 94.90 \\
spacy\_default\_lg & 85.28 & 71.19 & 76.97 & 60.62 & 92.90 & 94.84 \\
spacy\_bert       & 85.28 & 71.19 & 82.02 & 63.46 & 93.19 & 96.18 \\
\hline
gpt-4o                      & 91.70 & 93.09 & 85.99 & 87.47 & 96.01 & 98.01 \\
gpt-4o-mini-2024-07-18       & 90.25 & 93.96 & 83.91 & 84.40 & 95.97 & 97.41 \\
o3-mini-2025-01-31          & 94.28 & 95.38 & 88.39 & 90.59 & 96.98 & 97.06 \\
claude-3-5-sonnet-20241022  & 93.15 & 94.46 & 86.19 & 92.04 & 93.71 & 96.34 \\
claude-3-5-haiku-20241022   & 87.01 & 84.88 & 85.90 & 87.97 & 94.62 & 97.93 \\
GLM-4-0520                 & 88.44 & 87.72 & 84.53 & 81.54 & 94.77 & 97.51 \\
GLM-4-Long                 & 89.55 & 89.47 & 84.06 & 81.15 & 89.55 & 94.96 \\
\hline
\end{tabular}%
}
\caption{Segmentation, POS, and NER Accuracy for Poetry and Non-Poetry Texts.}
\label{tab:model_comparison}
\end{table*}

\subsection{Pre-1920 vs Post-1920}

Finally, when comparing texts from pre-1920 and post-1920, the models overall performed better when handling more contemporary data. Traditional NLP models showed noticeable improvements in word segmentation, nearly achieving a 90 percent F1 score across all models on post-1920 texts. This improvement reflects the recency bias of existing tools that are primarily trained on modern texts. The LLMs again exhibited a narrower performance gap between pre- and post-1920 data. Notably, gpt-4o's post-1920 results rival those of the computationally expensive o3-mini. Additionally, claude-3.5-sonnet yielded impressive and consistent results across all categories. While some of the average differences between pre-1920 and post-1920 are relatively minor, post-1920 results also had a much smaller interquartile range, representing more consistent performance from the various tools when working with post-1920 texts (see Figure 1). 

\begin{figure}
  \centering
  \includegraphics[width=\linewidth]{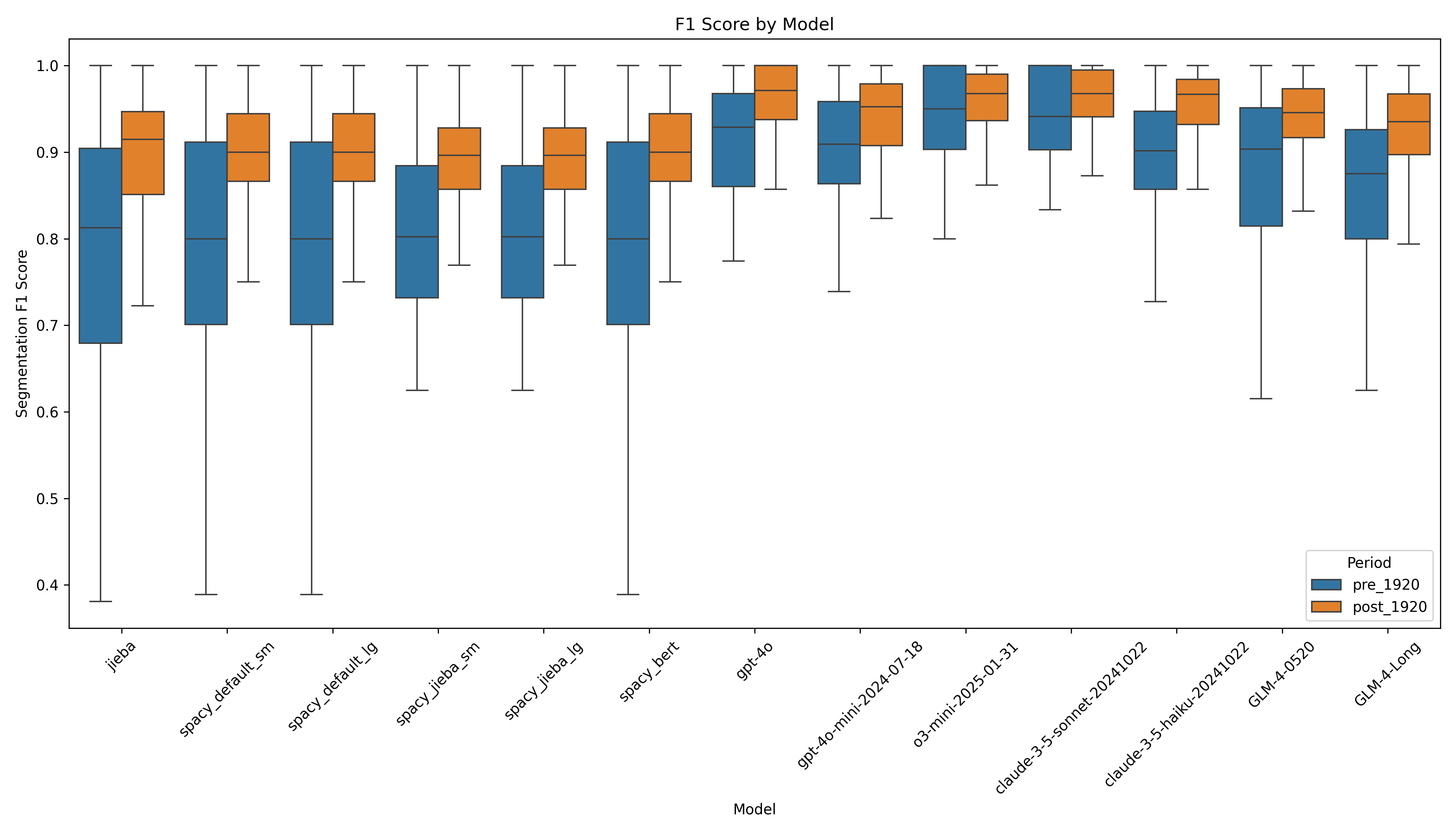}
  \caption{Boxplot of F1 Score for Temporal Change by Model for non-poetry texts, capturing the median (line), interquartile range (boxes), and spread of data (whiskers).}
\end{figure}

\begin{table*}[ht]
\centering
\resizebox{\textwidth}{!}{%
\begin{tabular}{lcccccc}
\hline
 & \multicolumn{2}{c}{\textbf{Seg\_F1 (\%)}} & \multicolumn{2}{c}{\textbf{POS\_Accuracy (\%)}} & \multicolumn{2}{c}{\textbf{NER\_Accuracy (\%)}} \\
\textbf{Model} & \textbf{Pre-1920} & \textbf{Post-1920} & \textbf{Pre-1920} & \textbf{Post-1920} & \textbf{Pre-1920} & \textbf{Post-1920} \\
\hline
jieba             & 76.30 & 88.42 & 52.20 & 44.03 & 89.11 & 94.89 \\
spacy\_jieba\_sm  & 77.75 & 88.13 & 64.40 & 72.94 & 86.72 & 93.99 \\
spacy\_jieba\_lg  & 77.75 & 88.13 & 71.74 & 78.43 & 90.40 & 93.80 \\
spacy\_default\_sm & 77.52 & 89.09 & 69.06 & 74.80 & 85.68 & 93.90 \\
spacy\_default\_lg & 77.52 & 89.09 & 74.47 & 78.19 & 91.07 & 93.79 \\
spacy\_bert       & 77.52 & 89.09 & 79.95 & 83.04 & 89.43 & 95.03 \\
\hline
gpt-4o                      & 85.67 & 94.66 & 83.42 & 87.25 & 94.49 & 96.75 \\
gpt-4o-mini-2024-07-18       & 87.38 & 91.66 & 86.88 & 82.44 & 93.96 & 96.97 \\
o3-mini-2025-01-31          & 91.50 & 95.65 & 87.98 & 88.59 & 96.49 & 97.23 \\
claude-3-5-sonnet-20241022  & 91.56 & 93.93 & 86.45 & 86.07 & 93.73 & 93.71 \\
claude-3-5-haiku-20241022   & 83.32 & 88.82 & 86.28 & 85.71 & 92.38 & 95.73 \\
GLM-4-0520                 & 85.55 & 89.86 & 84.80 & 84.40 & 94.06 & 95.13 \\
GLM-4-Long                 & 85.00 & 91.79 & 83.04 & 84.56 & 87.85 & 90.38 \\
\hline
\end{tabular}%
}
\caption{Segmentation, POS, and NER Accuracy for Pre- and Post-1920 Texts (non-poetry).}
\label{tab:model_comparison}
\end{table*}

\section{Discussion}

The impact of word segmentation choices on digital humanities (DH) and cultural analytics research, particularly for late 19th century to mid 20th century “transitional” Chinese texts, is significant and multifaceted. Proper segmentation enhances downstream tasks such as data analysis, pattern recognition, and cross-lingual/temporal studies. It improves the accuracy of frequency analyses, topic modeling, and semantic network analyses, while also making both transitional and classical Chinese texts more accessible.

While some have argued that word segmentation is becoming less relevant in NLP pipelines\cite{li2019word}, researchers in the humanities and social science still find it crucial. Character-based or sub-character methods (e.g., Byte Pair Encoding) often fall short for DH applications, where accurately representing search keywords and concepts is often prioritized over processing efficiency. Proper segmentation enables nuanced identification of linguistic patterns and cultural trends over time, facilitating comparative studies across languages and historical periods.

Domain-specific models like PKUSEG (which has been integrated into spaCy as its default tokenizer) offer improvements over generic tools, but to date have failed to curate training data for historical texts \cite{luo2019pkuseg}. LLMs show promise in overcoming these limitations through contextual learning. However, fine-tuning LLMs or adopting a hybrid approach to NLP tasks still requires manual engineering and domain expertise. While LLMs' pattern recognition capabilities for Chinese word segmentation, POS tagging, and NER are impressive, especially for corpora containing both modern and classical Chinese, prompt-engineered LLM tokenization can benefit from domain-specific knowledge and careful prompt design.
\begin{CJK}{UTF8}{bsmi}

Finally, it is important to note that without explicit word boundaries, there is often not a single \textit{correct} way to segment Chinese texts. Instead, word segmentation depends on interpretative choices that are shaped by both research objectives and historical context. In developing our ground truth dataset, we encountered several valid segmentation approaches. For instance, should 上海圖書館 (Shanghai Library) be treated as a single token, or should it be split into 上海 (Shanghai) and 圖書館 (Library)? Moreover, how we handle shifts in language might depend on our research questions. By the 1920s, the character pair 教授 (jiaoshou) should be seen as a single lexical item meaning “professor.” Conversely, in classical Chinese, these characters together meant “to impart knowledge,” with a two-token segmentation being more appropriate. However, researchers examining the semantic shift of jiaoshou from 1900 to 1950 might benefit from treating it consistently as a single token across time. Ultimately, the evolution of language and the inherent subjectivity in tokenization decisions underscore the complex nature of segmenting Chinese texts.
\end{CJK}
\section{Conclusion}
LLMs have demonstrated improved performance in handling complex Chinese language tasks, consistently outperforming traditional NLP tools across all metrics. LLMs also showed greater resilience in processing both poetic texts and language spanning multiple decades. These improvements over traditional tools like jieba and spaCy highlight the potential of LLMs in advancing Chinese NLP tasks. Further research should focus on optimizing LLMs to reduce computational costs while maintaining high accuracy, thereby making them more accessible for widespread use. Exploring hybrid models that combine the strengths of traditional NLP tools with LLMs could lead to more efficient and accurate systems for Chinese language processing and digital humanities applications.

\section*{Limitations}
Several notable limitations should be noted. First, our ground truth data is based on a relatively small sample of texts—we began with one hundred passages and later added one hundred more to test the robustness of our dataset. Although this augmentation did not change our overall findings, confirming our initial results, future studies would benefit from larger datasets to further validate the results. Additionally, we only evaluated out-of-the-box models rather than experimenting with fine-tuning or few-shot prompting. Future research could address these limitations by developing an open-source model that enhances scalability, efficiency, and broader accessibility.

\section*{Acknowledgements}
We would like to thank Liu Wei and his team at the Shanghai Library for their generous support in providing access to the data that made this project possible. 

\bibliography{main}
\bibliographystyle{acl_natbib}

\end{document}